\documentclass[11pt]{article}
\usepackage{acl}
\usepackage{times}
\usepackage{latexsym}
\usepackage[T1]{fontenc}
\usepackage[utf8]{inputenc}
\usepackage{inconsolata}
\usepackage{wrapfig} 
\usepackage{microtype}
\usepackage{hyperref}
\usepackage{url}
\usepackage{booktabs}
\usepackage{graphicx}
\usepackage{subcaption}
\usepackage{xspace}
\usepackage[most]{tcolorbox}              \usepackage{enumitem}
\newcommand{\env}{D3-Gym\xspace}
\usepackage{xcolor}
\usepackage{float}
\usepackage{listings}
\usepackage{fvextra}

\usepackage{makecell}

\usepackage{multirow}
\usepackage{soul}

\definecolor{darkblue}{rgb}{0, 0, 0.5}
\hypersetup{colorlinks=true, citecolor=darkblue, linkcolor=darkblue, urlcolor=darkblue}
\definecolor{SeaGreen}{HTML}{48c4a4}
\definecolor{OrangeRed}{HTML}{ff2c5c}
\definecolor{ForestGreen}{HTML}{209c54}

\newcommand{\hnm}[1]{{\color{black}#1}}

\title{\env: Constructing Real-World Verifiable Environments for Data-Driven Discovery}

\author{\textbf{Hanane Nour Moussa$^{1}$\thanks{Equal contribution.}, Yifei Li$^{1}$\footnotemark[1], Zhuoyang Li$^{1}$, Yankai Yang$^{2}$, Cheng Tang$^{1}$,} \\
\textbf{Tianshu Zhang$^{1}$, Nesreen K. Ahmed$^{3}$, Ali Payani$^{3}$, Ziru Chen$^{1}$, Huan Sun$^{1}$} \\
$^{1}$The Ohio State University \\
$^{2}$University of Wisconsin-Madison \\
$^{3}$Cisco Research \\
\texttt{\{moussa.45,li.14042,sun.397\}@osu.edu}
}

\begin{document}
\maketitle
\begin{abstract}
Despite recent progress in language models and agents for scientific data-driven discovery, advancing their capabilities is held back by the absence of verifiable environments representing real-world scientific tasks.
To fill this gap, we introduce \env, the first dataset of automatically constructed \emph{verifiable environments} for scientific \underline{D}ata-\underline{D}riven \underline{D}iscovery.
\env comprises 565 tasks from 239 real scientific repositories across four disciplines, each with a natural language instruction, an executable environment with pre-installed dependencies, dataset previews, a reference solution, and an \emph{automatically synthesized} evaluation script.
Our evaluation scripts achieve 87.5\% agreement with human-annotated gold standards and strong alignment in domain-specific evaluation logic.
Training on trajectories sampled from \env yields consistent gains across Qwen3 models on ScienceAgentBench, boosting Qwen3-32B SR@3 by 7.8 absolute points and shrinking the gap with strong proprietary models. \hnm{We further illustrate, through case studies, how \env environments can serve as a testbed for studying agentic optimization loops such as Autoresearch on real scientific workflows.}
We open-source \env, its creation workflow, sampled trajectories, and training scripts at \url{https://github.com/OSU-NLP-Group/D3-Gym}.
\end{abstract}

\section{Introduction}

Language agents \citep{su-etal-2024-language, sumers2024cognitive} are reshaping scientific research. Powered by large language models (LLMs), they can now meaningfully assist researchers in their workflows, from literature synthesis \citep{skarlinski2024languageagentsachievesuperhuman, asai2026synthesizing} to domain-specific scientific reasoning \citep{baker2025larchumanlevelconstrainedretrosynthesis, narayanan2025trainingscientificreasoningmodel}. In particular, there has been significant research interest in developing language agents for data-driven discovery \citep{hey2009fourthparadigm, majumder2024position}, where agents can programmatically test new hypotheses \citep{majumder2025discoverybench, mitchener2025bixbenchcomprehensivebenchmarkllmbased} and derive scientific insights from data \citep{chen2025scienceagentbench, shojaee2025llmsr, shojaee2025llmsrbenchnewbenchmarkscientific}. Despite this progress, further advancing these capabilities, particularly for open-weight LLMs \citep{zhu2025opensourcellmsstruggledata}, is held back by the absence of suitable infrastructure for scientific tasks.
\begin{figure}[t]
  \centering
  \vspace{-4mm}
  \includegraphics[width=0.48\textwidth]{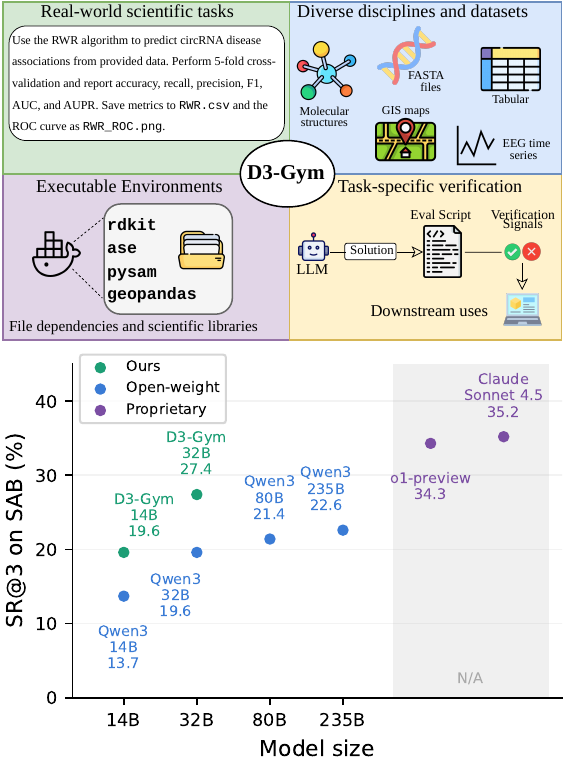}
  \caption{Top: Overview of \env. Bottom: Success Rate (SR@3) on ScienceAgentBench (SAB) \citep{chen2025scienceagentbench} of \env trained models based on Qwen3-14B and Qwen3-32B and other larger open-weight and proprietary reasoning models.}
  \label{fig:main_comparison}
  \vspace{-4mm}
\end{figure} 

A critical component in this infrastructure is \emph{real-world verifiable environments}. Recent work such as Autoresearch \citep{karpathy_autoresearch} highlights their value: given a manually prepared task instruction, a dataset, an evaluation script, and a repository with pre-installed dependencies, coding agents can use the executable feedback loop to improve performance and run dozens of experiments in a few hours. However, constructing such environments requires substantial manual effort, making them difficult to scale beyond a small number of tasks.

In this paper, we fill this gap with an automated pipeline that constructs real-world, verifiable coding environments for data-driven discovery. Unlike software engineering, where unit tests in repositories can be directly repurposed as verification signals \citep{yang2025swesmith, jain2025regym, pan2025training}, automatically constructing environments for scientific tasks is particularly challenging because related repositories rarely contain pre-existing evaluation scripts, and programs produce domain-specific outputs whose correctness cannot be assessed by any universal criterion. To address this, we start from a large pool of candidate tasks collected using AutoSDT \citep{autosdt} and design a series of rigorous filtering steps, followed by task execution and output validation, to retain high-quality, ecologically valid task instances. We then synthesize evaluation scripts in two stages: first writing a detailed, task-specific evaluation plan, and then translating that plan into an executable script to complete environment construction.

The resulting dataset, \env, comprises 565 verifiable environments from 239 scientific repositories in bioinformatics, computational chemistry, geographic information science, and psychology and cognitive neuroscience (\autoref{fig:main_comparison}). Each instance includes an executable environment with pre-installed dependencies, input data files and previews, a natural language task instruction, a reference solution adapted from the original repository, and a task-specific evaluation script with metrics and acceptance criteria. To our knowledge, \env is the first automatically constructed dataset with verifiable environments for scientific data-driven discovery.

Through comprehensive experiments, we demonstrate the quality of synthesized evaluation scripts and the utility of \env in two complementary settings. On 50 tasks with human-annotated gold evaluation scripts, our synthesized evaluation scripts achieve 87.5\% agreement on pass/fail verdicts and closely match the underlying evaluation logic. Furthermore, rejection sampling fine-tuning \citep{yuan2023scaling} on \env substantially improves Qwen3 models \citep{yang2025qwen3technicalreport} on ScienceAgentBench \citep{chen2025scienceagentbench}: our 32B model improves SR@3 by 7.8 absolute points, surpasses Qwen3-235B-A22B, and approaches strong proprietary models such as OpenAI o1 (\autoref{fig:main_comparison}). \hnm{Additionally, two case studies pairing a coding agent with \env environments in an Autoresearch-style search loop illustrate how the synthesized verification signals can support iterative agentic refinement and surface concrete behaviors of frontier coding agents on real scientific workflows.} Overall, \env provides a foundation for training, evaluation, and iterative refinement for data-driven discovery in real-world, verifiable environments.

\section{Constructing Verifiable Environments}
\label{sec:method}
\begin{figure*}[t]
    \centering
    \includegraphics[width=1\linewidth]{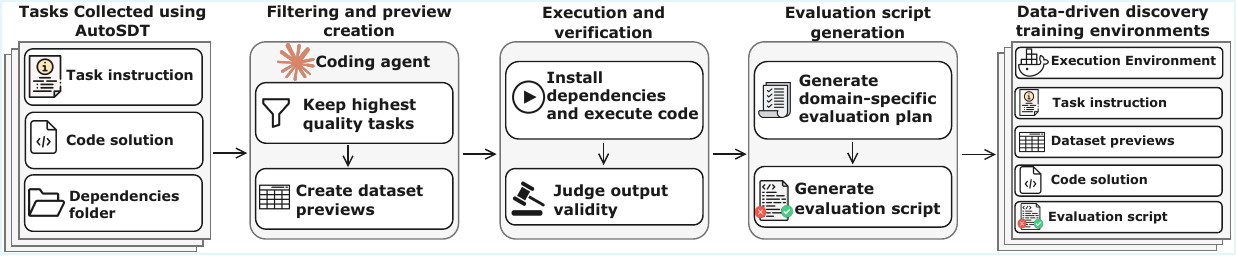}
    \caption{Overview of the \env construction workflow. Candidate tasks from scientific repositories \citep{autosdt} undergo quality filtering and dataset preview creation, then execution and validation to obtain reference outputs. These provide the context needed to generate high-quality, task-specific evaluation scripts via a planning-then-coding approach.}
    \label{fig:method}
    \vspace{-4mm}
\end{figure*}

As shown in Figure~\ref{fig:method}, we construct \env through a four-stage workflow. The first three stages (\S\ref{sec:task-preprocessing}) collect candidate tasks from scientific repositories, filter them for data integrity, and execute them to obtain reference outputs. These stages ensure task quality and produce the essential context that enables the core contribution of our workflow: automated generation of task-specific evaluation scripts with scientifically grounded metrics and acceptance criteria (\S\ref{sec:esg}). We validate these scripts against human-annotated gold standards in \S\ref{sec:quality_validation}. Further details are in \autoref{app:pipeline-details}.

\subsection{Task Pre-processing}
\label{sec:task-preprocessing}

\paragraph{Candidate Task Collection.} We collect an initial pool of candidate tasks using AutoSDT \citep{autosdt}, which crawls scientific repositories from GitHub, applies multi-step filtering to identify files implementing data-driven scientific workflows, adapts them into reference code solutions, and pairs each with a natural language task instruction. The pipeline also extracts a dependency folder for each task with all file dependencies from the original repository. Expert evaluation confirms that 93\% of collected tasks are scientifically meaningful \citep{autosdt}. We ensure that all repositories used in ScienceAgentBench \citep{chen2025scienceagentbench} are \textit{excluded}, as it is our evaluation set.

\paragraph{Filtering and Dataset Preview Creation.} A limitation of AutoSDT is that its code adaptation stage may substitute synthetic data when an input file is absent from the original repository \citep{autosdt}. Since our goal is to construct high-fidelity environments grounded in real scientific workflows, we discard such tasks with synthetic data and retain only those whose reference solutions operate exclusively on real data files in the original repository. Claude Code \citep{claudecode2025} verifies file dependencies and, for passing tasks, produces a \textit{dataset preview file} for each input data file: a structured excerpt exposing the data schema. These previews provide concrete data context and later inform evaluation script generation.

\paragraph{Execution and Output Verification.} We execute each remaining reference solution in an isolated environment, using \texttt{pipreqs}\footnote{https://pypi.org/project/pipreqs/} to install required packages. While the previous step ensures data integrity, execution can still fail or produce degenerate outputs due to environment mismatches or subtle errors from code adaptation. We employ a multimodal LLM-as-judge (GPT-5.2, 92.31\% agreement with human; Appendix~\ref{app:human-agreement}) to verify all requested outputs were produced and each output is meaningful and non-degenerate. Tasks failing either criterion are discarded. The verified outputs, together with the dataset previews, form the context for evaluation script generation.
\subsection{Evaluation Script Generation}
\label{sec:esg}
A central challenge is generating discriminative evaluation scripts that correctly capture what constitutes a correct solution for a given task while rejecting incorrect ones. Inspired by \citet{saha2025learning} and \citet{yang-etal-2025-code}, we decompose evaluation script generation into two phases: \textit{planning} and \textit{coding}. This separation encourages the planning LM to reason carefully about scientific validity without being distracted by implementation details, while allowing the coding LM to focus on producing correct, executable code faithful to the plan. This design choice is validated by our ablation in \S\ref{sec:quality_validation}, where removing the planning phase significantly reduces performance.

In the planning phase, we provide Claude Sonnet 4.5 with the task instruction, dataset previews, and verified reference outputs and leverage its parametric scientific knowledge to produce a detailed, task-specific evaluation plan. The plan specifies which output artifacts to inspect, which metrics are scientifically appropriate for each artifact, and the acceptance criteria, including tolerance thresholds, exact versus approximate matching, and domain-justified performance bounds. The evaluation plan is passed to Claude Sonnet 4.5 in a separate call, which generates the final executable evaluation script in a single pass. We refer to these automatically generated scripts as \textit{silver evaluation scripts}, to distinguish them from human-annotated \textit{gold evaluation scripts} used to assess their reliability.

\subsection{Validating Evaluation Script Quality}
\label{sec:quality_validation}

\begin{table*}[t]
\centering
\small
\begin{tabular}{@{}l ccc ccc@{}}
\toprule
& \multicolumn{3}{c}{\textbf{Pass/Fail Agreement (\%)}} & \multicolumn{3}{c}{\textbf{Evaluation Logic (1--5, $\uparrow$)}} \\
\cmidrule(lr){2-4} \cmidrule(lr){5-7}
\textbf{Method} & \textbf{Acc.} & \textbf{Spec.} & \textbf{Recall} & \textbf{Metric} & \textbf{Threshold} & \textbf{Artifact} \\
\midrule
Direct Prompting         & 86.1 & 100.0 & 0.0  & 2.07 & 2.16 & 2.66 \\
\midrule
Silver Eval. Scripts (Ours)                     & \textbf{87.5} & 91.0 & \textbf{66.1}& \textbf{4.00} & \textbf{3.82} & \textbf{4.20} \\
\quad w/o Planning        & 74.8 & 86.3 & 3.4 & 2.00 & 2.04 & 2.68 \\
\quad w/o Dataset Preview & 63.2 & 70.4 & 18.6 & 1.96 & 1.74 & 2.64 \\
\quad w/o Code Output     & 85.1 & 98.9 & 0.0& 2.89 & 2.64 & 2.62 \\
\bottomrule
\end{tabular}
\caption{Quality validation of silver evaluation scripts against human-annotated gold scripts. \textit{Left}: pass/fail agreement. High specificity with near-zero recall (e.g., Direct Prompting) indicates scripts that reject nearly all solutions indiscriminately. \textit{Right}: evaluation logic alignment on metric choice, threshold \& tolerance, and target artifact.}
\label{tab:quality_validation}
\vspace{-4mm}
\end{table*}
A critical property of any evaluation script is the reliability of its verification signal. We therefore rigorously evaluate our silver scripts along two complementary dimensions: \textit{(1) execution-based agreement}, which measures whether silver scripts produce the same pass/fail verdicts as human-annotated gold scripts, and \textit{(2) evaluation logic agreement}, which assesses whether the methodology aligns with that of gold scripts.

\paragraph{Setup.} We curate a held-out set of 50 task instances with gold evaluation scripts manually written and validated by Ph.D. students over 175 person-hours. For each, we generate a silver evaluation script using the method in \S\ref{sec:esg}. To obtain candidate solutions for execution-based metrics, we sample 10 solutions per task from Claude Opus 4.6 and GPT-5.4. These models were chosen because their higher execution success rates yield the largest pool of solutions that proceed to evaluation. After filtering for successful execution, this produces 424 candidate solutions, each scored by both the silver and gold scripts.

\paragraph{Metrics.} \textbf{(1) Accuracy} measures the proportion of candidate solutions for which the silver and gold scripts return the same verdict. \hnm{\textbf{(2) Specificity} measures the fraction of gold-failing solutions that also fail the silver script, the property that ensures incorrect solutions do not slip through verification into downstream use. \textbf{(3) Recall} measures the fraction of gold-passing solutions that also pass the silver script, capturing the rate at which correct solutions are retained.} Further, we assess silver scripts' scientific validity via an LLM-as-judge (85\% exact, 98\% within-1 agreement with human annotators; Appendix~\ref{app:human-agreement}) that scores each silver script against its gold counterpart on three aspects using a 1--5 Likert scale: \textbf{(4) Metric Choice}, \textbf{(5) Threshold \& Tolerance}, and \textbf{(6) Target Artifact}. Setup details are in \autoref{app:env-details}.

\paragraph{Baselines.} We compare our full method (\S\ref{sec:esg}) against a direct prompting baseline that provides the task instruction and code solution to an LLM and produces the evaluation script in one step. We also ablate three components: evaluation planning, dataset preview access, and code output access. All evaluation script generation methods use Claude Sonnet 4.5 as the backbone LLM.

\paragraph{Results.} As shown in Table~\ref{tab:quality_validation}, our pipeline is the only configuration that achieves substantial recall (66.1\%) while preserving high specificity (91.0\%). The baseline and all ablations show high specificity and low recall, indicating scripts that reject solutions indiscriminately. Removing evaluation planning reduces recall to just 3.4\%, removing the dataset preview introduces schema expectation errors, and removing access to code output reduces recall to 0\%. Our method consequently achieves the highest accuracy overall, with silver-accepted solutions reliably corresponding to those gold scripts would accept and a substantial fraction of correct solutions retained; the remaining disagreements concentrate at the gold pass/fail boundary, where solutions narrowly satisfy gold thresholds but fall outside the silver criteria (Appendix~\ref{app:es-examples}). The evaluation logic agreement scores further confirm silver scripts' methodological soundness: our method substantially outperforms the baseline and all ablations across metric choice, threshold \& tolerance, and target artifact.
\begin{figure*}[t]
    \centering
    \includegraphics[width=1\linewidth]{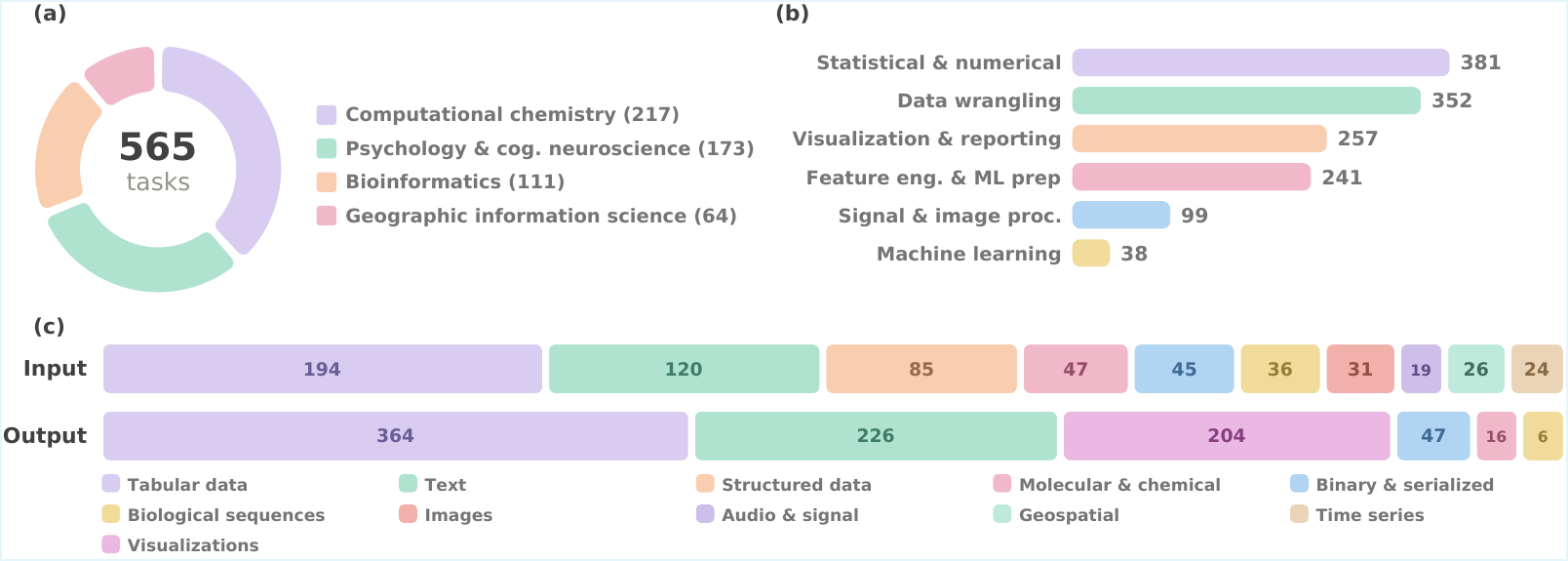}
    \caption{\env statistics. (a) Distribution of tasks by scientific discipline. (b) Distribution of tasks by type. Tasks carry multiple type labels (mean 2.4 labels per task). (c) Distribution of tasks by input and output modality. Tasks may carry multiple modalities (input: mean 1.11, max 3; output: mean 1.53, max 4).}
    \label{fig:stats}
\end{figure*}
\section{\env}
\label{sec:env}
Using our verifiable environment creation workflow (\S\ref{sec:method}), we present \env. Each task instance in \env comprises an executable environment with pre-installed dependencies, dataset preview files, a natural language task instruction, a scientist-authored reference solution, and a task-specific evaluation script generated automatically. \env comprises a diverse set of 565 tasks drawn from 239 scientific repositories. The synthesis of each task costs around \$3. We report the per-stage environments synthesis cost in Appendix~\ref{app:env-cost}.

\paragraph{Discipline, modality, and task type distribution.} As shown in \autoref{fig:stats}, \env spans four scientific disciplines: Computational Chemistry, Bioinformatics, Geographic Information Science, and Psychology \& Cognitive Neuroscience. Task types reflect the full arc of real-world scientific workflows, ranging from statistical analysis, numerical computation, and visualization to specialized tasks such as feature engineering, machine learning, and signal and image processing. \env also encompasses a wide spectrum of scientific data formats: in addition to conventional tabular, textual, and image data, many tasks operate on domain-specific representations such as molecular structure encodings (e.g., SMILES, FASTA), molecular simulation trajectories (e.g., TRAJ), and geospatial records (e.g., GPS logs). Outputs are similarly varied, including visualizations and serialized artifacts such as trained models, NumPy arrays, time series, maps, and domain-specific structures. We provide examples in Appendix~\ref{app:examples}.
\begin{table*}[t]
\centering
\small
\setlength{\tabcolsep}{4pt}
\resizebox{0.99\textwidth}{!}{%
\begin{tabular}{@{}l*{4}{c}*{4}{c}@{}}
\toprule
\multirow{2}{*}{\textbf{Method}} & \multicolumn{4}{c}{\textbf{ScienceAgentBench}} & \multicolumn{4}{c}{\textbf{ScienceAgentBench-Verified}} \\
\cmidrule(lr){2-5} \cmidrule(l){6-9}
& \textbf{SR} & \textbf{VER} & \textbf{SR@3} & \textbf{VER@3} & \textbf{SR} & \textbf{VER} & \textbf{SR@3} & \textbf{VER@3} \\
\midrule

\textit{Qwen3-4B-Instruct} \\
\quad Base
& 4.2 & 25.8 & 5.9 & 35.3
& 4.2 & 25.8 & 5.9 & 35.3 \\
\quad RFT-Distill
& 7.2 {\scriptsize\textcolor{green!50!black}{(+3.0)}}
& 30.4 {\scriptsize\textcolor{green!50!black}{(+4.6)}}
& 8.8 {\scriptsize\textcolor{green!50!black}{(+2.9)}}
& 47.1 {\scriptsize\textcolor{green!50!black}{(+11.8)}}
& 7.2 {\scriptsize\textcolor{green!50!black}{(+3.0)}}
& 30.4 {\scriptsize\textcolor{green!50!black}{(+4.6)}}
& 8.8 {\scriptsize\textcolor{green!50!black}{(+2.9)}}
& 47.1 {\scriptsize\textcolor{green!50!black}{(+11.8)}} \\
\addlinespace

\textit{Qwen3-4B} \\
\quad Base
& 4.9 & 19.6 & 6.9 & 27.5
& 5.2 & 20.1 & 7.3 & 29.2 \\
\quad RFT-Distill
& \textbf{7.2} & \textbf{23.5} & 8.8 & 37.3
& \textbf{7.2} & \textbf{23.2} & 8.8 & 36.3 \\
\quad RFT-Self
& 6.9 {\scriptsize\textcolor{green!50!black}{(+2.0)}}
& 23.2 {\scriptsize\textcolor{green!50!black}{(+3.6)}}
& \textbf{11.8} {\scriptsize\textcolor{green!50!black}{(+4.9)}}
& \textbf{40.2} {\scriptsize\textcolor{green!50!black}{(+12.7)}}
& 6.9 {\scriptsize\textcolor{green!50!black}{(+1.7)}}
& \textbf{23.2} {\scriptsize\textcolor{green!50!black}{(+3.1)}}
& \textbf{11.8} {\scriptsize\textcolor{green!50!black}{(+4.5)}}
& \textbf{38.2} {\scriptsize\textcolor{green!50!black}{(+9.0)}} \\
\addlinespace

\textit{Qwen3-8B} \\
\quad Base
& 6.5 & 29.7 & 11.8 & 43.1
& 8.2 & 31.2 & \textbf{13.8} & \textbf{44.7} \\
\quad RFT-Distill
& 8.5 & 31.7 & \textbf{14.7} & \textbf{48.0}
& 8.2 & 29.1 & 13.7 & 43.1 \\
\quad RFT-Self
& \textbf{10.1} {\scriptsize\textcolor{green!50!black}{(+3.6)}}
& \textbf{36.9} {\scriptsize\textcolor{green!50!black}{(+7.2)}}
& 12.8 {\scriptsize\textcolor{green!50!black}{(+1.0)}}
& 43.1 {\scriptsize(0.0)}
& \textbf{10.8} {\scriptsize\textcolor{green!50!black}{(+2.6)}}
& \textbf{33.7} {\scriptsize\textcolor{green!50!black}{(+2.5)}}
& 13.7 {\scriptsize\textcolor{red!70!black}{(-0.1)}}
& 41.2 {\scriptsize\textcolor{red!70!black}{(-3.5)}} \\
\addlinespace

\textit{Qwen3-14B} \\
\quad Base
& 9.1 & 35.3 & 13.7 & 49.0
& 10.3 & 36.3 & 15.0 & 50.0 \\
\quad RFT-Distill
& 10.5 & 35.0 & 13.7 & 48.1
& 10.5 & 33.0 & 13.7 & 47.1 \\
\quad RFT-Self
& \textbf{13.1} {\scriptsize\textcolor{green!50!black}{(+4.0)}}
& \textbf{37.6} {\scriptsize\textcolor{green!50!black}{(+2.3)}}
& \textbf{19.6} {\scriptsize\textcolor{green!50!black}{(+5.9)}}
& \textbf{54.9} {\scriptsize\textcolor{green!50!black}{(+5.9)}}
& \textbf{14.1} {\scriptsize\textcolor{green!50!black}{(+3.8)}}
& \textbf{37.3} {\scriptsize\textcolor{green!50!black}{(+1.0)}}
& \textbf{20.6} {\scriptsize\textcolor{green!50!black}{(+5.6)}}
& \textbf{54.9} {\scriptsize\textcolor{green!50!black}{(+4.9)}} \\
\addlinespace

\textit{Qwen3-32B} \\
\quad Base
& 14.7 & 36.3 & 19.6 & 50.9
& 15.0 & 37.0 & 20.0 & 51.0 \\
\quad RFT-Self
& \textbf{18.9} {\scriptsize\textcolor{green!50!black}{(+4.2)}}
& \textbf{43.8} {\scriptsize\textcolor{green!50!black}{(+7.5)}}
& \textbf{27.4} {\scriptsize\textcolor{green!50!black}{(+7.8)}}
& \textbf{58.8} {\scriptsize\textcolor{green!50!black}{(+7.9)}}
& \textbf{19.3} {\scriptsize\textcolor{green!50!black}{(+4.3)}}
& \textbf{41.8} {\scriptsize\textcolor{green!50!black}{(+4.8)}}
& \textbf{28.4} {\scriptsize\textcolor{green!50!black}{(+8.4)}}
& \textbf{54.9} {\scriptsize\textcolor{green!50!black}{(+3.9)}} \\
\bottomrule
\end{tabular}%
}
\caption{Main results on ScienceAgentBench and ScienceAgentBench-Verified (\%, $\uparrow$). We report average Success Rate (SR) and Valid Execution Rate (VER) across three independent runs, as well as best-of-three performance (SR@3 and VER@3). For RFT-Self, numbers in parentheses indicate absolute change from the corresponding base model. Best results within each model block are bolded.}
\label{tab:main_results_combined}
\end{table*}
\paragraph{Scientific package coverage and task difficulty.} Tasks draw on both the general-purpose scientific stack (\texttt{numpy}, \texttt{pandas}, \texttt{matplotlib}, \texttt{scipy}) and domain-specific libraries including \texttt{rdkit}, \texttt{ase}, \texttt{Bio}, \texttt{pysam}, \texttt{LFPy}, \texttt{dyconnmap}, \texttt{geopandas}, and \texttt{folium}. This breadth of domain-specific tooling reflects \env's genuine scientific diversity and is a key source of its difficulty. To gauge task difficulty, we sample 75 tasks and evaluate GPT-5.4, Claude Opus 4.6, Qwen3-235B-A22B, and Qwen3-32B (Appendix~\ref{app:task-difficulty}). Even the most capable frontier models solve only around a third of the tasks (GPT 5.4: 38.66\%), and none consistently produce fully executable code, with GPT-5.4 and Claude Opus 4.6 achieving only 82.66\% and 78.66\% valid execution rates. Overall, these results confirm our tasks are solvable yet non-trivial.

\section{Training using \env}
\label{sec:training}

\env provides a versatile training environment that supports various approaches for improving model capabilities. Here, we demonstrate its utility by training open-weight models for scientific data-driven discovery---a setting where reliable, transparent, and reproducible systems are especially important \citep{spirling2023opensource}. Given the current performance gap of open-weight models on these tasks, we use training approaches that provide dense supervision and are effective at expanding model capabilities \citep{yue2025does}. We therefore adopt rejection-sampling fine-tuning (RFT) \citep{yuan2023scaling}, where a model generates trajectories with full reasoning traces and final solutions in a training environment, retaining only successful ones for fine-tuning. We experiment with \textbf{RFT-Distill}, where a smaller student learns from a stronger teacher, and \textbf{RFT-Self}, where a model learns from its own successful trajectories.
\subsection{Experimental Setup}
\paragraph{Models.} We use four student models from the Qwen3 family \citep{yang2025qwen3technicalreport}: Qwen3-4B, Qwen3-8B, and Qwen3-14B in thinking mode, and Qwen3-4B-Instruct as a non-thinking variant. For expert distillation, we use Qwen3-32B as the teacher.
\paragraph{Training.} For each model, we sample 16 trajectories per task in \env, yielding 6{,}780 trajectories per model, retaining only those that pass the silver evaluation scripts. The number of successful trajectories reflects growing capability with model size: Qwen3-4B (1{,}126), Qwen3-8B (1{,}306), Qwen3-14B (1{,}788), and Qwen3-32B (2{,}153). In \textbf{RFT-Distill}, students are fine-tuned on successful trajectories from the Qwen3-32B teacher, and in \textbf{RFT-Self}, models are fine-tuned on their own successful trajectories. For Qwen3-4B-Instruct, we apply only RFT-Distill using the final solutions without reasoning traces.

\paragraph{Evaluation.} We evaluate on \textbf{ScienceAgentBench} \citep{chen2025scienceagentbench}, where a model must generate a Python program that processes input data, implements the required analysis, and saves results to the correct output path. We also evaluate on \textbf{ScienceAgentBench-Verified}, a manually verified version introduced here, where we fixed nine unclear instructions, two erroneous gold programs, and one redundant gold execution result (Appendix~\ref{app:sab_verified}). We report \textbf{Success Rate (SR)}, whether outputs satisfy human-annotated success criteria, and \textbf{Valid Execution Rate (VER)}, whether the program executes without errors and produces output in the correct location. For each model we conduct 3 independent inference runs and report both average and best-of-3 performance. Further training details are in \autoref{app:training-details}.

\subsection{Training Results}
\paragraph{Training on \env consistently improves performance across model sizes.} As shown in \autoref{tab:main_results_combined}, training on trajectories from \env improves both SR and VER across all model sizes, indicating models learn to produce programs that are both executable and scientifically correct. Gains are evident not only in average performance but also in best-of-3 performance (SR@3), reflecting broader capability gains rather than only improved typical-case performance. These gains scale with model size: SR@3 improves by +1.0, +5.9, and +7.8 for the 8B, 14B, and 32B models, respectively.

\begin{figure*}[t]
    \centering
    \begin{subfigure}{0.49\linewidth}
        \centering
        \includegraphics[width=\linewidth]{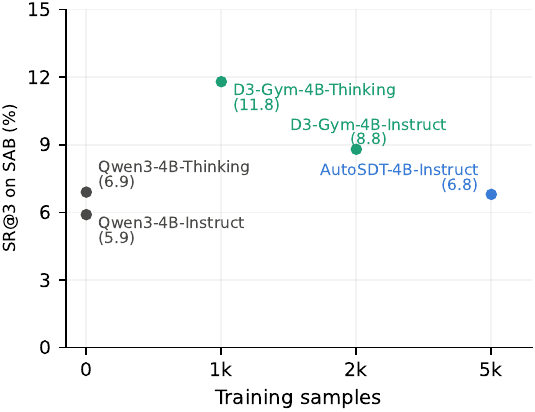}
        \caption{}
        \label{fig:sr3_sab}
    \end{subfigure}
    \hfill
    \begin{subfigure}{0.49\linewidth}
        \centering
        \includegraphics[width=\linewidth]{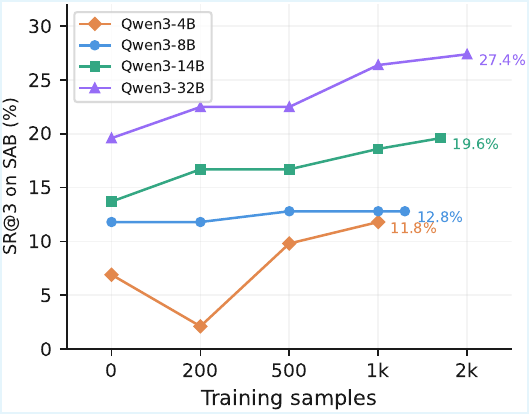}
        \caption{}
        \label{fig:scaling}
    \end{subfigure}
    \caption{(a) Training on AutoSDT-5K. (b) RFT-Self scaling across models.}
    \label{fig:analysis}
\end{figure*}

\paragraph{Models learn effectively from their own trajectories.} Across models, RFT-Self generally matches or outperforms RFT-Distill, with the gap widening at larger scales. For Qwen3-4B and Qwen3-8B, neither approach is consistently better. At 14B, however, RFT-Self becomes consistently superior, improving SR@3 by 5.9 points while RFT-Distill yields no improvement, even degrading VER@3. A plausible explanation is that on-policy trajectories are better aligned with the model's own distribution \citep{zhao2026self}. To verify this, we score self-generated and 32B teacher-generated responses under each RFT-Self model on 64 randomly sampled prompts, using negative log-likelihood (NLL) and perplexity (PPL); lower values mean the response is more likely under that model's own distribution. Both Qwen3-8B (NLL: 0.252 vs.\ 0.652; PPL: 1.29 vs.\ 1.93) and Qwen3-14B (NLL: 0.261 vs.\ 0.543; PPL: 1.30 vs.\ 1.73) assign substantially lower values to self-generated responses, which are preferred on all prompts in both cases. This supports that self-generated trajectories are closer to the model's learned distribution, enabling more effective learning.

\paragraph{\env-32B approaches the performance of larger and proprietary models.} As shown in \autoref{fig:main_comparison}, training on \env enables smaller models to match or surpass significantly larger ones. Notably, \env-32B outperforms both Qwen3-80B-A3B and Qwen3-235B-A22B, two larger mixture-of-experts reasoning models, and narrows the gap with the strong proprietary o1-preview \citep{openai2024o1} and Claude Sonnet 4.5 \citep{anthropic2025sonnet45}. Similarly, \env-14B sees substantial gains, matching Qwen3-32B despite being less than half its size.

\subsection{Analysis}

\paragraph{Training on \env outperforms static SFT data.} \autoref{fig:sr3_sab} compares training on \env data against AutoSDT-5K \citep{autosdt}, a static dataset of 5K instruction--solution pairs that lacks executable environments or verification signals. When used for standard SFT, \env is more sample-efficient: Qwen3-4B-Instruct trained on \env solutions achieves roughly 30\% relative improvement over its AutoSDT-5K baseline using less than half the training samples, likely due to the rigorous filtering that guarantees task quality in \env. Beyond sample efficiency, \env's executable environment and verification signals enable sampling full reasoning trajectories, unlocking thinking model training; Qwen3-4B improves from 6.9\% to 11.8\% SR@3 when trained on its own verified reasoning trajectories.

\begin{figure*}[t]
    \centering
    \begin{subfigure}{0.49\linewidth}
        \centering
        \includegraphics[width=\linewidth]{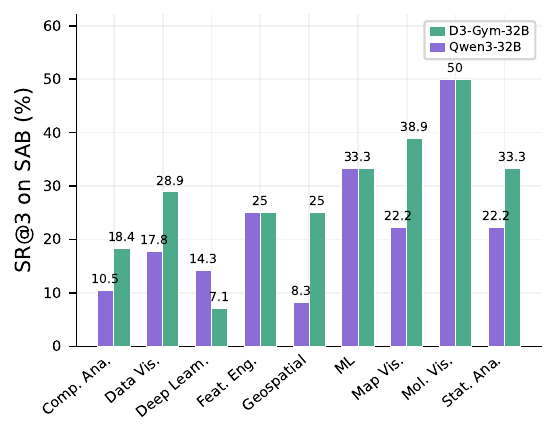}
        \caption{}
        \label{fig:per_category}
    \end{subfigure}
    \hfill
    \begin{subfigure}{0.49\linewidth}
        \centering
        \includegraphics[width=\linewidth]{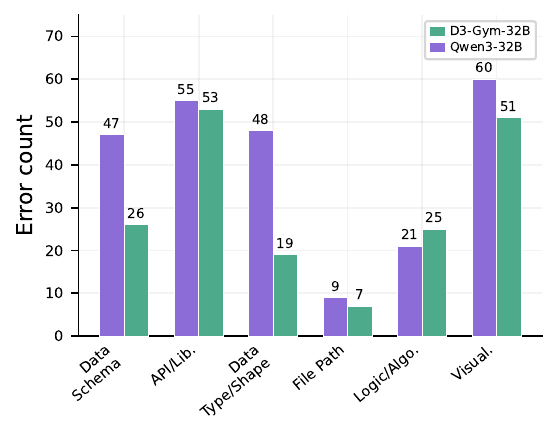}
        \caption{}
        \label{fig:error_analysis}
    \end{subfigure}
    \caption{(a) Per-category SR@3 on ScienceAgentBench. (b) Error-type breakdown for Qwen3-32B vs. \env-32B.}
    \label{fig:analysis2}
\end{figure*}

\paragraph{Larger models show more consistent improvement with training scale.} \autoref{fig:scaling} shows SR@3 under RFT-Self as training trajectories increase. The 14B and 32B models improve steadily with additional data, whereas the 8B model gains marginally before plateauing and the 4B model initially degrades with few training samples before recovering. This suggests larger models better extract useful signal from increasing training data, consistent with previous studies \citep{autosdt, jain2025regym}.

\paragraph{Performance improves across diverse task categories.} \autoref{fig:per_category} breaks down SR@3 by task category for the Qwen3-32B base model and \env-32B. Training yields gains across many categories including computational analysis, data and map visualization, geospatial analysis, and statistical analysis. This breadth suggests \env tasks are sufficiently diverse to expand model capabilities across scientific domains rather than overfitting to a narrow subset of skills.

\paragraph{Failure modes are greatly mitigated, but some persist.} We manually analyze errors from both models across six types: \textit{Data schema}, \textit{API \& library misuse}, \textit{Data type \& shape}, \textit{File path}, \textit{Logical \& algorithmic}, and \textit{Visualization}. As shown in \autoref{fig:error_analysis}, \env-32B reduces data schema errors by 44.7\%, data type/shape errors by 60.4\%, and visualization errors by 15.0\%, improving execution validity. Logical and algorithmic errors show a slight increase: with more programs passing execution, the model more frequently produces runnable but semantically incorrect code. API \& library errors also remain prevalent despite a modest decrease, suggesting a deep limitation in current LMs' parametric knowledge.
\section{Autoresearch with \env Environments}
\label{sec:autoresearch}
Beyond training, \env environments can also support studying automated research with coding agents: many data-driven discovery tasks can be framed as optimization problems, where an agent repeatedly proposes solutions and improves them against an executable evaluation signal. Unlike the training experiments in \S\ref{sec:training}, where silver evaluation scripts act as binary filters over sampled trajectories, here we expose their raw score and study whether it is reliable enough to guide a frontier coding agent across many iterations.

\paragraph{Setup.} We adopt the Autoresearch harness \citep{karpathy_autoresearch} with modifications to adapt it to Codex \citep{openai2025codex} and enable persistent memory across sessions, and run Codex with GPT-5.4 for 40 iterations on two \env tasks. The first is a spatial interpolation problem: given temperature readings from 73 German weather stations, predict the temperature on a regular grid covering the whole country. The prescribed method is universal kriging \citep{cressie1993statistics}, essentially Gaussian-process regression with a trend term as its mean function, here a north--south temperature gradient. Solutions are scored by the RMSE of their temperature map against the reference solution's (lower is better; original pass threshold RMSE $<1.5$). The second is more open-ended: train a graph neural network with Chebyshev-polynomial spectral filters \citep{defferrard2016convolutional} to predict 12 quantum-chemical properties of small molecules, scored by MAE averaged over the properties (threshold MAE $<2.0$). Iteration 1 is the agent's single-shot solution and serves as the baseline. Task specifications and harness details are in Appendix~\ref{app:autoresearch}.
\begin{figure*}[t]
    \centering
    \begin{subfigure}{0.49\linewidth}
        \centering
        \includegraphics[width=\linewidth]{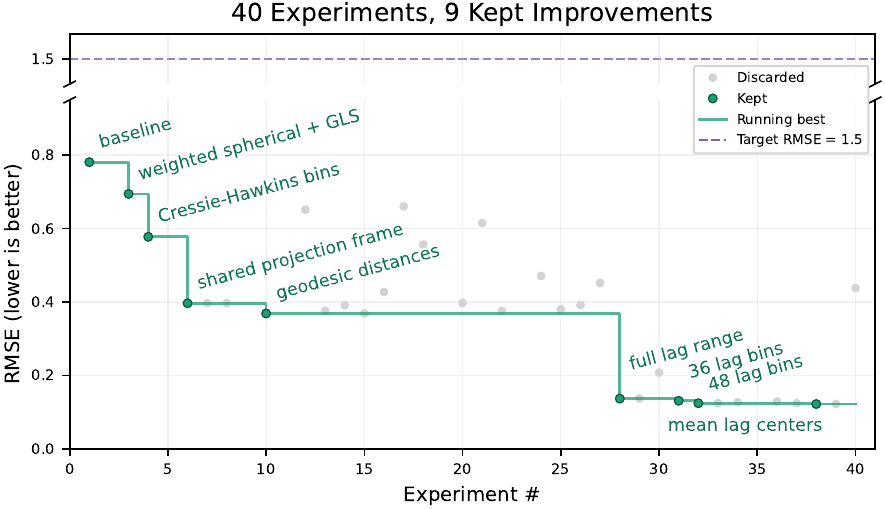}
        \caption{Spatial Interpolation of German Temperature Data.}
        \label{fig:autoresearch_temp}
    \end{subfigure}
    \hfill
    \begin{subfigure}{0.49\linewidth}
        \centering
        \includegraphics[width=\linewidth]{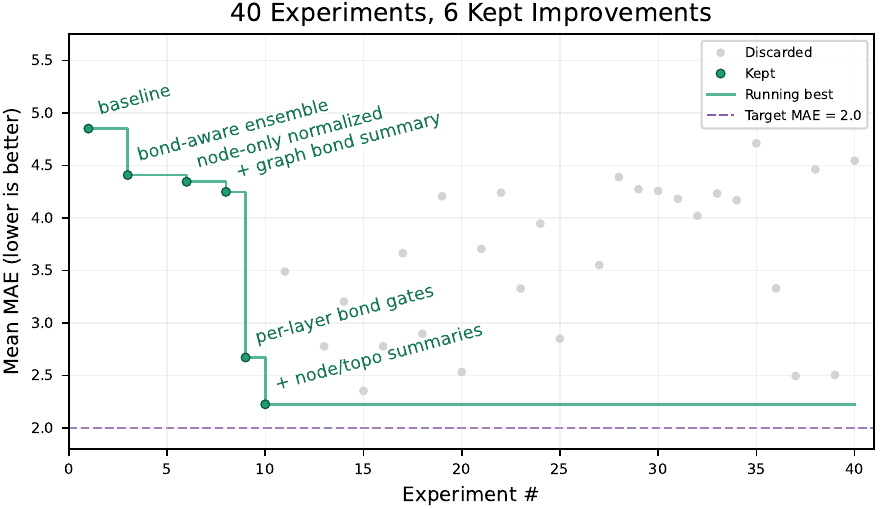}
        \caption{Molecular Property Prediction with ChebyNet.}
        \label{fig:autoresearch_chebynet}
    \end{subfigure}
    \caption{Autoresearch optimization trajectories on two \env tasks over 40 iterations.}
    \label{fig:autoresearch}
\end{figure*}

\paragraph{Spatial interpolation task.} As shown in \autoref{fig:autoresearch_temp}, the baseline (RMSE $0.7806$) already satisfies the threshold, yet the raw RMSE remains informative and continues to guide improvement. The agent never departs from the prescribed method; every change it keeps refines how the model is fit to the data: a better estimate of how temperature similarity decays with distance, fitting that is less sensitive to outlying stations, consistent geographic coordinates with distances measured along the Earth's surface, and progressively finer discretization of the station distances (the figure labels each change). The score falls from $0.7806$ to $0.1221$, a $6.4\times$ reduction, showing that the evaluation scripts synthesized by \env are reliable enough to serve as a stable optimization objective.

\paragraph{ChebyNet task.} Here the agent must make modeling, representation, and training decisions rather than refine a fixed procedure. Its early gains come from neural-network engineering: ensembling, richer inputs (bond information, molecule-level summary features), feature normalization, and gating messages by bond type. MAE falls from $4.8505$ to $2.2249$, a $2.2\times$ reduction, but then saturates short of the threshold. The cause is a systematic failure: the training set is small, so the reference solution trains a very small model for a single epoch, whereas the agent keeps scaling the model up and training it for hundreds of epochs. This overfitting is never diagnosed or corrected; the agent continues to add architectural complexity even as the score plateaus.

Together, the case studies show that silver evaluation scripts give agents reliable signal to improve substantially over their own single-shot solutions, while open-ended tasks surface failure modes such as overfitting and weak self-diagnosis. Automatically constructed verifiable environments can therefore serve not only as training infrastructure but also as testbeds for studying how frontier coding agents search, improve, and fail on realistic scientific workflows.
\section{Related Work}
\label{app:related-work}
\paragraph{Environment creation for coding tasks.}
Recent work generates executable training environments for coding tasks. In software engineering, SWE-Gym \citep{pan2025training}, RepoST \citep{xie2025repost}, R2E-Gym \citep{jain2025regym}, and SWE-smith \citep{yang2025swesmith} build environments from real GitHub repositories, leveraging commits and unit tests for task synthesis and verification. Beyond software engineering, MLE-Dojo \citep{qiang2025mledojo}, MLGym \citep{nathani2025mlgym}, and MLE-Smith \citep{qiang2026mlesmith} target machine learning tasks where correctness is assessed via standard metrics like F1 and RMSE. Scientific repositories offer neither: codebases rarely include test suites, and correctness is inseparable from the scientific domain, varying with each task's data modality and analytical goal. Evaluation logic must therefore be constructed from scratch per task.

\paragraph{Environments for data-driven discovery.}
Most closely related are recent efforts targeting data-driven discovery. \citet{autosdt} introduce a pipeline for collecting scientific coding tasks at scale to construct SFT datasets, but the AutoSDT dataset lacks the execution environments and evaluation scripts needed for trajectory generation or solution validation. We build on their task collection pipeline and address this limitation by constructing executable, verifiable environments around the collected tasks. \hnm{\citet{nie2026dsgym} introduce DSGym, drawing tasks from existing benchmarks, bioinformatics queries, and Kaggle competitions, with evaluation inherited from these sources. \env differs in both sourcing and verification: tasks are drawn directly from real scientific repositories across four disciplines, and evaluation scripts are automatically synthesized.}

\paragraph{Evaluation script generation.}
Recent works tackle automated test and evaluation generation for general coding, both to improve test quality \citep{he2025hardtestssynthesizinghighqualitytest, liu2025rstarcoder} and to enable downstream training \citep{zeng-etal-2025-acecoder, wang-etal-2025-codecontests, prasad2025learning}. There, correctness largely reduces to string or near-string comparison against a reference. Our setting differs fundamentally: evaluation logic for scientific data-driven discovery is tightly coupled to the task's domain, analytical goal, and data modality, requiring reasoning about which metrics, thresholds, and output artifacts are scientifically appropriate.
\section{Conclusion}
We introduce \env, the first automatically constructed dataset of real-world verifiable environments for scientific data-driven discovery, and show its support for both model training and iterative agent refinement. \env opens a path toward scientific agents that can test, improve, and learn in realistic research settings.
\section*{Limitations}
\label{app:limitations}
We recognize the following limitations and future work directions:
\paragraph{\env Environment Scale.} \env currently comprises 565 tasks, which is limited compared to training environments in other domains such as software engineering. However, this reflects the difficulty of sourcing high-quality, scientifically meaningful tasks rather than a constraint of our workflow, which can readily collect more tasks provided sufficient repositories are available. Moreover, \env's executable and verifiable nature enables sampling thousands of training trajectories, which our results confirm is sufficient to yield substantial improvements across model sizes.
\paragraph{Training method.} We focus on RFT for training, which provides dense supervision that is particularly effective when base model performance is weak, as is the case for the open-weight models we train. Our results confirm the effectiveness of this approach in expanding model capabilities across all scales, and we further show that \env supports training-free agent refinement. While our verifiable environments would also support other training techniques including reinforcement learning, models' limited baseline capabilities, the complexity of data-driven discovery tasks, and the sparsity of the reward signal make this a challenging direction that we leave for future work.
\paragraph{Evaluation.} We evaluate the models trained on \env on ScienceAgentBench \citep{chen2025scienceagentbench}, as it is the most suitable benchmark that matches \env's target capability, namely producing executable programs for real-repository scientific data analysis and judging them against expert-annotated success criteria. We regard evaluation across a wider set of benchmarks, as suitable ones emerge in this space, as an important direction for future work.
\section*{Ethical Considerations}
We make a concerted effort to ensure \env tasks are adapted from publicly available scientific repositories with permissive licenses. A complete per-repository license breakdown is in Appendix~\ref{app:licenses}. 
\section*{Author Contributions}
HNM led the project, implemented the D3-Gym creation workflow, conducted quality evaluation, model training, and analysis, and wrote the manuscript. YL co-led the project, managed the curation of the validation set for quality evaluation, conducted training and evaluation experiments, contributed to the development of SAB-verified, and assisted with manuscript writing. ZL assisted with initial candidate task collection and co-managed the annotation of the validation set. YY, CT, and TZ contributed to the annotation of the validation set. NKA and AP provided feedback during biweekly project discussions. ZC provided guidance on project direction and ideas, contributed to the development of SAB-verified, explored Autoresearch with \env, and revised the manuscript. HS advised the project and provided guidance, contributed to core ideas, and revised the manuscript.
\section*{Acknowledgments}
The authors would like to thank colleagues from the OSU NLP group and Amazon AGI team for constructive feedback.
The authors thank Xinming Tu for reaching out and helping to construct ScienceAgentBench-Verified.
This research was sponsored in part by NSF OAC 2112606, Cisco, and Ohio Supercomputer Center \citep{OhioSupercomputerCenter1987}.
The views and conclusions contained herein are those of the authors and should not be interpreted as representing the official policies, either expressed or implied, of the U.S. government. The U.S. Government is authorized to reproduce and distribute reprints for Government purposes notwithstanding any copyright notice herein.
\bibliography{custom}

\section*{Appendix}
We provide more details omitted from the main text in the Appendix as follows:
\begin{itemize}
    \item Appendix \ref{app:pipeline-details}: Details of \env construction workflow
        \begin{itemize}
            \item Appendix \ref{app:pipeline-stats}: Pipeline Statistics
            \item Appendix \ref{app:prompts}: Prompts
            \item Appendix \ref{app:human-agreement}: LLM judge agreement with human
        \end{itemize}
    \item Appendix \ref{app:env-details}: \env details
        \begin{itemize}
            \item Appendix \ref{app:examples}: Example tasks
            \item Appendix \ref{app:env-cost}: Cost breakdown
            \item Appendix \ref{app:quality-validation-details}: Quality validation of \env verification signal
            \item Appendix \ref{app:es-examples}: Examples of Gold and Silver Evaluation Scripts
            \item Appendix \ref{app:task-difficulty}: Task difficulty
        \end{itemize}
    \item Appendix \ref{app:training-details}: Training Details
    \item Appendix \ref{app:sab_verified}: ScienceAgentBench-Verified
    \item Appendix \ref{app:autoresearch}: Autoresearch Experimental Details
    \item Appendix \ref{app:licenses}: Repository Licenses
    \item Appendix \ref{app:ai_use}: Disclosure of AI use
\end{itemize}
\setcounter{table}{0}
\renewcommand\thetable{\Alph{section}.\arabic{table}}
\setcounter{figure}{0}
\renewcommand\thefigure{\Alph{section}.\arabic{figure}}
\appendix
\section{Details of \env construction workflow}
\label{app:pipeline-details}
\subsection{Pipeline Statistics}
\label{app:pipeline-stats}

The creation workflow of \env follows the steps provided in \autoref{sec:method}. Concretely, we begin by collecting 5{,}111 candidate tasks using the AutoSDT pipeline \citep{autosdt}. After filtering to retain only tasks whose reference solutions operate exclusively on dependencies present in the original repository and generating dataset previews, 1{,}586 tasks remain. Of these, 1{,}263 execute successfully in isolated conda environments and produce output artifacts. Finally, 565 pass the multimodal LLM-as-judge output verification and receive silver evaluation scripts, constituting the final \env environment --- approximately 11\% of the initial pool, reflecting the stringent filtering needed to obtain the highest quality tasks. 
\subsection{Prompts}
\label{app:prompts}
In this section we present all the agent and LLM instructions used in our workflow, namely for task filtering and preview creation, output validity judgment, and evaluation script planning and generation. 
\begin{tcolorbox}[
    enhanced,
    breakable,
    colback=blue!3!white,
    colframe=blue!50!black,
    title={\textbf{Task Filtering Agent Prompt}},
    fonttitle=\bfseries\small,
    fontupper=\scriptsize,
    left=4pt, right=4pt, top=4pt, bottom=4pt,
    boxrule=0.5pt
  ]
  
  You are a coding agent with access to the local file system. You are given a candidate task consisting of a Python program (\texttt{program.py}) and its associated data files. Your objective is twofold: (1)~determine whether this task uses only real data and real library dependencies, with no mock, stub, or simulated logic; and (2)~if the task is valid, generate dataset preview files for downstream use.

  \medskip
  \textbf{Phase 1: Task Validation.} Carry out the following steps to determine task validity:

  \begin{enumerate}[leftmargin=*,itemsep=2pt]
      \item \textbf{Identify Data Files.} Inspect the Python program and extract every file path that it attempts to open, read, load, or process. For each path:
      \begin{itemize}[itemsep=1pt]
          \item Verify whether the file exists on the current file system.
          \item Read a small portion of the file to confirm it contains real, meaningful data (not empty, not placeholder, not all zeros, not constant meaningless values, not randomly generated toy content).
      \end{itemize}
      If all referenced data files exist and contain valid data, set \texttt{dummy\_data = 0}; otherwise set \texttt{dummy\_data = 1}.

      \item \textbf{Check for Mock or Simulated Logic.} Analyze the program for any of the following:
      \begin{itemize}[itemsep=1pt]
          \item Mock objects or \texttt{unittest.mock} usage (e.g., \texttt{Mock()}, \texttt{MagicMock}, \texttt{patch}).
          \item Custom stub or fake class implementations replacing real libraries.
          \item Synthetic data generation used as primary input instead of loading from files.
          \item Hardcoded dummy values pretending to be real data.
          \item \texttt{try-except} blocks catching \texttt{ImportError} and defining mock replacements.
          \item Any workaround for missing dependencies that bypasses the real library.
      \end{itemize}
      If the program contains any such logic, set \texttt{has\_mock = 1}; otherwise set \texttt{has\_mock = 0}.

      \item \textbf{Render Verdict.} If \texttt{dummy\_data == 0} and \texttt{has\_mock == 0}, set \texttt{valid = 1}; otherwise set \texttt{valid = 0}. If \texttt{valid = 0}, skip Phase~2 and proceed directly to the output.
  \end{enumerate}

  \medskip
  \textbf{Phase 2: Dataset Preview Generation} (only if \texttt{valid = 1}).\\
  For each data file referenced by the program, generate a preview file showing only the raw data schema:
  \begin{itemize}[leftmargin=*,itemsep=1pt]
      \item For CSV or tabular data: the header row plus 3--5 example rows.
      \item For JSON: a snippet showing the structure with 1--2 entries.
      \item For text files: the first few lines of raw content.
      \item For binary or image files: a brief factual description (e.g., ``PNG images, 224$\times$224, RGB'').
  \end{itemize}
  Each preview must follow the format:
  \begin{verbatim}
[START Preview of <file_path>]
<raw data snippet>
[END Preview of <file_path>]
  \end{verbatim}


  \end{tcolorbox}
  
\begin{tcolorbox}[
  enhanced,
  breakable,
  colback=blue!3!white,
  colframe=blue!50!black,
  title={\textbf{Multimodal Gold-Result Judge Prompt}},
  fonttitle=\bfseries\small,
  fontupper=\scriptsize,
  left=4pt, right=4pt, top=4pt, bottom=4pt,
  boxrule=0.5pt
]

You are a meticulous evaluator for data-driven discovery workflows. Each task represents an analysis workflow on a dataset---such as statistical analysis, machine learning model training, data visualization, or computational simulation.

\medskip
\textbf{Inputs Provided.}
\begin{enumerate}[leftmargin=*,itemsep=2pt]
    \item A \emph{task instruction} describing the analysis to perform.
    \item \emph{Dataset preview} files showing the structure and content of the input data.
    \item A list of output files created in \texttt{pred\_results/} and their content (a.k.a gold results).
\end{enumerate}

The gold results were generated by synthesized programs. Complete correctness is \emph{not} guaranteed due to possible errors in code synthesis, data processing, or the execution environment.

\medskip
\textbf{Evaluation Criteria.} For a task to \textsc{pass}, \emph{all} of the following must hold:

\begin{enumerate}[leftmargin=*,itemsep=2pt]
    \item \textbf{All requested output files must exist.} Read the task instruction to identify every output file that should be produced. Verify each one appears in the output files listing. If any requested file is missing, the task \textsc{fails}.

    \item \textbf{Each output file must be valid.} For every file, verify that:
    \begin{itemize}[itemsep=1pt]
        \item It is non-empty and contains substantive content (no placeholders, no ``TODO'', no all-zeros).
        \item It contains no error messages or stack traces.
        \item Its content aligns with what the task instruction requested.
        \item Its format and values are reasonable for the analysis type.
    \end{itemize}
    Use your domain knowledge to assess whether the results are plausible (e.g., metric values within expected ranges, predictions that are sensible given the data).
\end{enumerate}

\medskip
\textbf{Evaluation Procedure.}
\begin{enumerate}[leftmargin=*,itemsep=2pt]
    \item \textbf{Identify required output files} from the task instruction.
    \item \textbf{Check file existence} against the list of required outputs.
    \item \textbf{Validate each file} for correctness, completeness, and validity.
    \item \textbf{Render a final decision}: \textsc{pass} only if all files exist and all are valid.
\end{enumerate}

\medskip
\textbf{Output Format.} Return a JSON object with exactly three fields:

{\ttfamily
\{"valid": true|false, "reason": "<explanation>"\}}
\end{tcolorbox}

\begin{tcolorbox}[
  enhanced,
  breakable,
  colback=blue!3!white,
  colframe=blue!50!black,
  title={\textbf{Evaluation Script Planner Prompt}},
  fonttitle=\bfseries\small,
  fontupper=\scriptsize,
  left=4pt, right=4pt, top=4pt, bottom=4pt,
  boxrule=0.5pt
]

You are an expert in scientific data analysis and evaluation methodology. Your task is to analyze a data-driven discovery task and produce a concrete evaluation plan that a coding agent will implement.

\medskip
\textbf{Inputs Provided.}
\begin{itemize}[leftmargin=*,itemsep=2pt]
    \item \textbf{Task Description}: the natural-language instruction defining the analysis to perform.
    \item \textbf{Dataset Information}: the dataset path, folder structure, and schema previews.
    \item \textbf{Expected Output Files}: the list of files the gold program produces.
    \item \textbf{Reference Result Files}: the actual content of the reference outputs.
\end{itemize}

\medskip
\textbf{Instructions.} Produce a structured evaluation plan covering the following:

\begin{enumerate}[leftmargin=*,itemsep=2pt]
    \item \textbf{Task Type.} Classify the task (e.g., classification, regression, clustering, visualization, statistical analysis, simulation).

    \item \textbf{Evaluation Metrics.} Specify which metric(s) are appropriate and justify why:
    \begin{itemize}[itemsep=1pt]
        \item \emph{Classification}: if the dataset is imbalanced, prefer F1-score, balanced accuracy, AUROC, or AUPRC over accuracy.
        \item \emph{Regression}: consider MAE, RMSE, $R^2$, or domain-specific error metrics.
        \item \emph{Statistical analysis}: consider $p$-values, effect sizes, confidence intervals.
        \item \emph{Clustering}: consider silhouette score, Davies--Bouldin index, or domain-specific quality measures.
        \item \emph{Other}: specify domain-appropriate criteria.
    \end{itemize}

    \item \textbf{Success Thresholds.} Define specific threshold values for each metric, grounded in domain standards and task complexity.

    \item \textbf{Special Considerations.} Note any domain-specific requirements such as handling of missing data, biological versus statistical significance, cross-validation needs, or output format constraints.

    \item \textbf{Evaluation Steps.} Provide a clear 3--5 step evaluation procedure that the coding agent should follow.
\end{enumerate}

\medskip
\textbf{Output Requirements.} The plan must be concise, unambiguous, and specify exactly one evaluation strategy. Avoid presenting multiple alternatives or optional branches---the coding agent will implement the plan directly as written.

\end{tcolorbox}

\begin{tcolorbox}[
  enhanced,
  breakable,
  colback=blue!3!white,
  colframe=blue!50!black,
  title={\textbf{Evaluation Script Coder Prompt}},
  fonttitle=\bfseries\small,
  fontupper=\scriptsize,
  left=4pt, right=4pt, top=4pt, bottom=4pt,
  boxrule=0.5pt
]

You are a Python coding assistant. Your task is to generate an evaluation script for a scientific computing benchmark, following the evaluation plan provided below.

\medskip
\textbf{Inputs Provided.}
\begin{itemize}[leftmargin=*,itemsep=2pt]
    \item \textbf{Evaluation Plan}: the structured plan produced by the evaluation planner.
    \item \textbf{Task Instruction}: the natural-language description of the analysis task.
    \item \textbf{Dataset Information}: the dataset path, folder structure, and schema previews.
    \item \textbf{Expected Output Files}: the list of files the gold program produces in \texttt{pred\_results/}.
    \item \textbf{Reference Result Files}: the reference output contents in \texttt{gold\_results/}.
\end{itemize}

\medskip
\textbf{Structure.} The generated script must adhere to the following:

\begin{enumerate}[leftmargin=*,itemsep=2pt]
    \item \textbf{Directory layout (hardcoded paths):}
    \begin{itemize}[itemsep=1pt]
        \item Predicted results: \texttt{./pred\_results/}
        \item Reference results: \texttt{./reference\_results/}
    \end{itemize}

    \item \textbf{Function signature:} Define a top-level \texttt{eval()} function taking no parameters. It returns a tuple \texttt{(result, message)} where \texttt{result} is a boolean (\texttt{True}/\texttt{False}) indicating pass/fail and \texttt{message} is a string with details.

    \item \textbf{Visual evaluation} (for tasks producing plots or images): use an LLM-as-a-judge approach to compare the predicted and reference visual outputs.

    \item \textbf{Main block:}
    \begin{verbatim}
if __name__ == "__main__":
    ok, msg = eval()
    print(ok, msg)
    \end{verbatim}

    \item \textbf{Error handling:} Wrap the body of \texttt{eval()} in a \texttt{try/except} so that unexpected errors return \texttt{(False, f"Error: \{e\}")} rather than crashing.

    \item \textbf{File existence checks:} Verify that both predicted and gold files exist before loading. Return \texttt{(False, "Missing file: ...")} if any file is absent.
\end{enumerate}

\medskip
\textbf{Output Format.} Respond with \emph{only} the Python source code for the evaluation script. Do not include any explanation or markdown formatting.

\end{tcolorbox}

\subsection{LLM Judge Agreement with Human}
\label{app:human-agreement}

To validate the reliability of the LLM-as-judge used for output verification (\S\ref{sec:method}), one author independently annotated 52 task outputs and we measured agreement with the LLM judge (GPT-5.2).

Table~\ref{tab:judge-agreement} reports the results. The LLM judge achieves 92.31\% raw agreement with the human annotator (48/52) and a Cohen's $\kappa$ of 0.85, corresponding to significant agreement. Treating the human labels as ground truth, the judge attains perfect precision (1.00) and 0.87 recall, with an F1 of 0.93. All four disagreements are false negatives, i.e. cases where the LLM rejected outputs that the human accepted, indicating that the judge is conservative rather than permissive. This bias is desirable in our setting, as the judge may discard a small number of valid tasks but does not allow low-quality outputs into \env tasks.
\begin{table}[h]
\centering
\small
\begin{tabular}{@{}lc@{}}
\toprule
\textbf{Metric} & \textbf{Value} \\
\midrule
Raw Agreement     & 92.31\% \\
Cohen's $\kappa$  & 0.85 \\
Precision         & 1.00 \\
Recall            & 0.87 \\
F1                & 0.93 \\
\bottomrule
\end{tabular}
\caption{Agreement between the LLM judge (GPT-5.2) and a human annotator on output verification for 52 task instances. Precision, recall, and F1 treat human labels as ground truth.}
\label{tab:judge-agreement}
\end{table}

\section{\env Details}
\label{app:env-details}

\subsection{Example Tasks}
\label{app:examples}
In this section we present examples of tasks from \env belonging to the disciplines of geographic information science and computational chemistry. The examples include the task instruction, dataset previews, and excerpts from the reference solution and evaluation script code. 


\begin{tcolorbox}[
  enhanced,
  breakable,
  colback=blue!3!white,
  colframe=blue!50!black,
  title={\textbf{Example Task: Geographic Information Science}},
  fonttitle=\bfseries\small,
  fontupper=\scriptsize,
  left=4pt, right=4pt, top=4pt, bottom=4pt,
  boxrule=0.5pt
]
\label{ex:kriging}

\textbf{Task Instruction.}
Perform spatial interpolation of German temperature data using Universal Kriging with a north-south drift model.
Use the temperature observation data from \texttt{temp\_obs.txt} and German border coordinates from \texttt{de\_borders.txt}.
First, estimate the empirical variogram of the temperature measurements and fit a Spherical variogram model to characterize the spatial correlation structure.
Then apply Universal Kriging to interpolate temperature values across a regular lat-lon grid covering Germany ($47$--$56.1^{\circ}$N, $5$--$16.1^{\circ}$E with $0.1^{\circ}$ resolution), incorporating a linear north-south drift function based on latitude.
Save the fitted variogram plot, the interpolated temperature field and mean drift field as CSV files, temperature field visualizations, a north-south cross-section plot, and the variogram model parameters.

\medskip
\textbf{Dataset Preview.}
\begin{itemize}[leftmargin=*,itemsep=2pt]
  \item \texttt{temp\_obs.txt} --- 73 DWD weather stations (id, latitude, longitude, temperature in $^{\circ}$C):
\begin{verbatim}
  # id,  lat,      lon,     temp
  44   52.9336   8.2370   15.70
  73   48.6159  13.0506   13.90
  91   50.7446   9.3450   17.00
  96   52.9437  12.8518   21.90
\end{verbatim}
  \item \texttt{de\_borders.txt} --- German border as lon/lat coordinate pairs:
\begin{verbatim}
  9.52402  47.5242
  9.35000  47.5989
  9.18281  47.6707
\end{verbatim}
\end{itemize}

\medskip
\textbf{Reference Solution (excerpt).}
\begin{verbatim}
  import gstools as gs, numpy as np

  data = np.loadtxt("benchmark/.../temp_obs.txt")
  ids, lat, lon, temp = data.T

  # Empirical variogram + Spherical model fit
  bin_center, vario = gs.vario_estimate(
      (lat, lon), temp, latlon=True,
      geo_scale=gs.KM_SCALE, max_dist=900)
  model = gs.Spherical(latlon=True, geo_scale=gs.KM_SCALE)
  model.fit_variogram(bin_center, vario, nugget=False)

  # Universal Kriging with latitude drift
  def north_south_drift(lat, lon): return lat
  uk = gs.krige.Universal(model=model, cond_pos=(lat, lon),
      cond_val=temp, drift_functions=north_south_drift)

  g_lat = np.arange(47, 56.1, 0.1)  # 92 points
  g_lon = np.arange(5, 16.1, 0.1)   # 112 points
  uk.set_pos((g_lat, g_lon), mesh_type="structured")
  uk(return_var=False, store="temp_field")
  uk(only_mean=True, store="mean_field")
\end{verbatim}

\medskip
\textbf{Evaluation Script (excerpt).}
The script validates the variogram model type and fitted parameters, computes RMSE and $R^2$ between the predicted temperature field and the gold standard, and checks physical validity of the interpolated values including the expected latitudinal temperature gradient.
\begin{verbatim}
  # Grid dimensions must be exactly (92, 112)
  expected_rows = len(np.arange(47, 56.1, 0.1))
  expected_cols = len(np.arange(5, 16.1, 0.1))
  assert pred_temp.shape == (expected_rows, expected_cols)

  # Variogram: must be Spherical; validate fitted parameters
  assert "spherical" in model_text.lower()
  assert variance > 0 and nugget >= 0 and nugget < variance
  assert 10 < len_scale < 5000  # km (gold ~ 596 km)

  # Temperature field RMSE vs gold < 1.5; R^2 vs gold > 0.60
  temp_rmse = np.sqrt(np.mean((pred_temp - gold_temp)**2))
  ss_res = np.sum((pred_temp - gold_temp)**2)
  ss_tot = np.sum((gold_temp - np.mean(gold_temp))**2)
  temp_r2 = 1.0 - ss_res / ss_tot
  assert temp_rmse < 1.5 and temp_r2 > 0.60

  # Physical plausibility: temps in [-10, 40] C
  assert np.min(pred_temp) >= -10 and np.max(pred_temp) <= 40

  # North-south gradient: |T_south - T_north| > 0.5 C
  south = np.mean(pred_mean[:10, :])   # lat ~47-48
  north = np.mean(pred_mean[-10:, :])  # lat ~55-56
  assert abs(south - north) > 0.5
\end{verbatim}

\end{tcolorbox}


\begin{tcolorbox}[
  enhanced,
  breakable,
  colback=blue!3!white,
  colframe=blue!50!black,
  title={\textbf{Example Task: Computational Chemistry}},
  fonttitle=\bfseries\small,
  fontupper=\scriptsize,
  left=4pt, right=4pt, top=4pt, bottom=4pt,
  boxrule=0.5pt
]
\label{ex:madelung}

\textbf{Task Instruction.}
Calculate the Madelung constants for a series of ionic crystal structures using the Ewald summation method. Process the crystal structure files \texttt{NaCl.vasp}, \texttt{CsCl.vasp}, \texttt{ZnO-Hex.vasp}, \texttt{ZnO-Cub.vasp}, \texttt{TiO2.vasp}, and \texttt{CaF2.vasp} to determine the electrostatic potential at the reference atom site in each crystal lattice. The Madelung constant quantifies the electrostatic interactions in ionic crystals by summing contributions from both real space and reciprocal space using the Ewald summation technique with appropriate convergence parameters. Save the computed results to a formatted table showing the crystal name, reference atom symbol, and calculated Madelung constant value for each structure.

\medskip
\textbf{Dataset Preview.}
\begin{itemize}[leftmargin=*,itemsep=2pt]
  \item Six VASP POSCAR crystal structure files spanning different crystal symmetries:

  \texttt{NaCl.vasp} --- rock salt (face-centered cubic):
\begin{verbatim}
  Na Cl
   1.0000000000000000
       0.000000    2.820000    2.820000
       2.820000    0.000000    2.820000
       2.820000    2.820000    0.000000
   Na  Cl
     1   1
  Cartesian
    0.000000  0.000000  0.000000
    2.820000  0.000000  0.000000
\end{verbatim}

  \texttt{TiO2.vasp} --- rutile (tetragonal, 6 atoms):
\begin{verbatim}
  Ti O
   1.0000000000000000
       4.593700    0.000000    0.000000
       0.000000    4.593700    0.000000
       0.000000    0.000000    2.958700
   Ti  O
     2   4
  Cartesian
    0.000000  0.000000  0.000000
    2.296850  2.296850  1.479350
    1.400160  1.400160  0.000000
    3.193540  3.193540  0.000000
    3.697010  0.896690  1.479350
    0.896690  3.697010  1.479350
\end{verbatim}
\end{itemize}

\medskip
\textbf{Reference Solution (excerpt).}
\begin{Verbatim}[fontsize=\scriptsize, breaklines, breakanywhere, commandchars=\\\{\}]

  from ase.io import read
  import numpy as np

  ZZ = {'Na': 1, 'Cl': -1, 'Cs': 1, 'Ca': 2, 'F': -1,
        'Ti': 4, 'Zn': 2, 'O': -2}

  class EwaldSum:
      def __init__(self, atoms, charges):
          self.cell = atoms.get_cell()
          self.positions = atoms.get_positions()
          self.charges = [charges[s] for s in           
          atoms.get_chemical_symbols()]

      def get_madelung(self, eta=None):
          # Optimal eta from cell volume
          volume = abs(np.linalg.det(self.cell))
          eta = (np.pi * len(self.atoms) / volume)**(1/3)
          # Sum real-space + reciprocal-space + self-correction
          potential = self._real_space(eta) + self._recip_space(eta)
          potential -= 2 * eta / np.sqrt(np.pi) * self.charges[0]
          return -potential / self.charges[0]

  for crys in ['NaCl', 'CsCl', 'ZnO-Hex', 'ZnO-Cub', 'TiO2', 'CaF2']:
      atoms = read(f"benchmark/.../{crys}.vasp")
      M = EwaldSum(atoms, ZZ).get_madelung()
\end{Verbatim}

\medskip
\textbf{Evaluation Script (excerpt).}
The script compares computed Madelung constants against established literature reference values for each crystal system (e.g., NaCl $= 1.7476$, CsCl $= 1.7627$, TiO$_2$ $= 2.408$), checking that all six crystals are present, values are positive, and the relative percentage error stays below 5\%.
\begin{Verbatim}[fontsize=\scriptsize, breaklines, breakanywhere, commandchars=\\\{\}]
  # Literature reference values for each crystal system
  REFERENCE = {
      'NaCl':    {'value': 1.7476, 'tolerance': 0.05},
      'CsCl':    {'value': 1.7627, 'tolerance': 0.05},
      'ZnO-Hex': {'value': 1.6413, 'tolerance': 0.10},
      'ZnO-Cub': {'value': 1.6381, 'tolerance': 0.10},
      'TiO2':    {'value': 2.408,  'tolerance': 0.15},
      'CaF2':    {'value': 2.5194, 'tolerance': 0.10},
  }

  # All 6 crystals must be present
  for crystal in ['NaCl','CsCl','ZnO-Hex','ZnO-Cub','TiO2','CaF2']:
      assert crystal in pred_results

  # Values must be positive; RPE < 5% vs literature
  for crystal, (atom, pred_val) in pred_results.items():
      assert pred_val > 0
      ref = REFERENCE[crystal]['value']
      rpe = abs(pred_val - ref) / ref * 100.0
      assert rpe < 5.0
\end{Verbatim}

\end{tcolorbox}
\subsection{Cost Breakdown}
\label{app:env-cost}
\autoref{tab:cost} breaks down the cost of each stage of the workflow for constructing \env. On average, the cost of constructing one environment in \env is around \$3. Most of this cost is incurred by the candidate task collection stage using the AutoSDT pipeline \cite{autosdt} due to its multi-step search and filtering for scientifically meaningful tasks across thousands of repository files. This cost remains within reasonable bounds and is far lower than that of recent work synthesizing training environments for software engineering tasks \cite{fu2026davincienvopensweenvironment}.
\begin{table}[t]
\centering
\small
\begin{tabular}{@{}lc@{}}
\toprule
\textbf{Workflow Stage} & \textbf{Cost (\$)} \\
\midrule
Candidate Task Collection & 1{,}094.5 \\
Filtering \& Dataset Preview Creation & 404.5 \\
Code Execution & 0.0 \\
Output Validation & 46.8 \\
Evaluation Planning \& Generation & 153.3 \\
\midrule
\textbf{Total} & \textbf{1{,}699.1} \\
\bottomrule
\end{tabular}
\caption{Cost breakdown (\$) by workflow stage for constructing \env.}
\label{tab:cost}
\end{table}
\subsection{Quality Validation of \env Verification Signal}
\label{app:quality-validation-details}
\subsubsection{LLM Judge Prompt}
\label{app:judge-prompt}
\begin{tcolorbox}[
    enhanced,
    breakable,
    colback=blue!3!white,
    colframe=blue!50!black,
    title={\textbf{Evaluation Logic Agreement Judge Prompt}},
    fonttitle=\bfseries\small,
    fontupper=\scriptsize,
    left=4pt, right=4pt, top=4pt, bottom=4pt,
    boxrule=0.5pt
  ]

  You are a scientific computing evaluation expert. You will be given two evaluation scripts written for the same data analysis task: a \emph{gold} (human-written) script and a \emph{silver} (LLM-generated) script. Your job is to rigorously assess whether the silver script makes the same scientific evaluation decisions as the gold. The gold script serves as the authoritative reference; the silver script is being evaluated for alignment with it.

  \medskip
  Focus on the substance of the evaluation logic---the scientific properties being measured, the pass/fail criteria, and the artifacts being compared---not on superficial differences such as code style, variable naming conventions, or choice of library. Your assessment should reflect whether the two scripts would reach the same pass/fail conclusion on a correctly implemented solution.

  \medskip
  Score each of the following three aspects \emph{independently} on a 1--5 Likert scale:
  \begin{itemize}[leftmargin=*,itemsep=1pt]
      \item \textbf{1 -- Completely missing or wrong}: the aspect is absent from the silver script, or the silver script's implementation is entirely incorrect with respect to the gold.
      \item \textbf{2 -- Present but fundamentally different}: the silver script addresses the aspect, but in a way that is scientifically incompatible with the gold (e.g., evaluating a different property altogether).
      \item \textbf{3 -- Partially aligned}: the silver script captures the intent of the gold, but contains differences substantial enough to change pass/fail outcomes on some inputs.
      \item \textbf{4 -- Well aligned}: the silver script is consistent with the gold, with only minor differences unlikely to affect pass/fail outcomes in practice.
      \item \textbf{5 -- Fully aligned or equivalent}: the silver script implements the same evaluation logic as the gold, or a functionally equivalent alternative.
  \end{itemize}

  \medskip
  \textbf{Aspect 1: Evaluation Metric.} Determine whether the silver script measures the same scientific property as the gold. Consider the metric definition, its mathematical formulation, and the correctness of the implementation. Two scripts that compute the same quantity using different libraries or formulations are aligned. However, an incorrect implementation of the intended metric (e.g., computing concordance without accounting for censoring in survival analysis, or using accuracy on a heavily imbalanced classification task where the gold uses F1) constitutes a misalignment and should be scored accordingly.

  \medskip
  \textbf{Aspect 2: Acceptance Criteria.} For each pass/fail threshold defined in the gold script, determine whether the silver script applies a corresponding threshold on the same or an equivalent metric. Consider whether the thresholds are calibrated such that a correct solution passing the gold script would also pass the silver, and vice versa. Absent thresholds, dramatically different threshold values, or thresholds applied to the wrong metric all indicate misalignment.

  \medskip
  \textbf{Aspect 3: Target Artifact.} Determine whether the silver script reads the correct output files, extracts the correct columns, keys, or fields from those files, and loads ground-truth data from the correct reference source. Misalignment includes reading the wrong file, parsing incorrect columns, or comparing against the wrong ground-truth artifact.

  \medskip
  \textbf{Output Format.} Return only a JSON object:

  {\ttfamily
  \{\\
  \quad "evaluation\_metric": \{"score": <1-5>, "rationale": "<one sentence>"\},\\
  \quad "acceptance\_criteria": \{"score": <1-5>, "rationale": "<one sentence>"\},\\
  \quad "target\_artifact": \{"score": <1-5>, "rationale": "<one sentence>"\}\\
  \}}
  \end{tcolorbox}

\subsubsection{LLM Judge - Human agreement}
To validate the LLM-as-judge used for evaluating silver script quality (\S\ref{sec:quality_validation}), one author independently scored 20 silver evaluation scripts against their gold counterparts on the same three aspects, i.e. metric choice, threshold \& tolerance, and target artifact, using the 1--5 Likert scale in \S\ref{app:judge-prompt}. Table~\ref{tab:inter-annotator-agreement} reports agreement with the LLM judge (Claude Sonnet 4.5).

Across all three dimensions, the judge achieves 85\% exact agreement and 98\% within-one agreement with the human annotator. Agreement is highest on metric choice (90\% exact) and lowest on threshold \& tolerance (80\% exact), though even in the latter case all scores fall within one point of the human rating. 

\begin{table}[h]
\centering
\small
\begin{tabular}{@{}lcc@{}}
\toprule
\textbf{Aspect} & \textbf{Exact Agr.} & \textbf{Within-1 Agr.} \\
\midrule
Metric Choice          & 0.90 & 0.95 \\
Threshold \& Tolerance & 0.80 & 1.00 \\
Target Artifact        & 0.85 & 1.00 \\
\midrule
\textbf{Overall}       & 0.85 & 0.98 \\
\bottomrule
\end{tabular}
\caption{Agreement between the LLM judge (Claude Sonnet 4.5) and a human annotator on metric choice, threshold \& tolerance, and target artifact.}
\label{tab:inter-annotator-agreement}
\end{table}
\subsection{Examples of Gold and Silver Evaluation Scripts}
\label{app:es-examples}

\begin{tcolorbox}[
  enhanced,
  breakable,
  colback=blue!3!white,
  colframe=blue!50!black,
  title={\textbf{Example 1}},
  fonttitle=\bfseries\small,
  fontupper=\scriptsize,
  left=4pt, right=4pt, top=4pt, bottom=4pt,
  boxrule=0.5pt
]
\label{ex:eval-delaney}

\textbf{Task.}
Train a model on the Delaney solubility dataset and produce point
predictions~(\texttt{y\_pred}) together with calibrated per-molecule
uncertainty estimates~(\texttt{y\_std}).

\medskip
\textbf{Key differences.}
The gold script checks RMSE~$\le 1.75$ and whether
$|\text{error}_i| \le 2\sigma_i$ for $\ge$70\% of molecules.
The silver script tightens RMSE to~1.70 and switches to the
standard 95\% CI convention ($\hat{y} \pm 1.96\sigma$), requiring
$\ge$72\% coverage.
For well-calibrated uncertainties the two coverage definitions are
nearly equivalent, but the silver thresholds are marginally tighter
on both axes.

\medskip
\textbf{Gold evaluation script.}
Uses a $2\sigma$ envelope for uncertainty calibration and an RMSE
threshold of~1.75.
\begin{Verbatim}[fontsize=\scriptsize, breaklines, breakanywhere, commandchars=\\\{\}]
  # --- thresholds ---
  RMSE_THRESHOLD = 1.75                          # <-- slightly more lenient
  COVERAGE_THRESHOLD = 0.70

  # --- RMSE ---
  rmse = float(np.sqrt(np.mean((y_pred - y_true) ** 2)))

  # --- uncertainty calibration: 2-sigma envelope ---
  abs_error = np.abs(y_pred - y_true)
  coverage_2sigma = float(np.mean(              # <-- P(|error| <= 2 * std)
      abs_error <= 2.0 * y_std))

  if rmse > RMSE_THRESHOLD:
      return False, f"RMSE too high: {rmse:.4f}"
  if coverage_2sigma < COVERAGE_THRESHOLD:
      return False, f"P(|error| <= 2*sigma)={coverage_2sigma:.3f}"
\end{Verbatim}

\medskip
\textbf{Silver evaluation script.}
Uses the standard 95\% confidence-interval convention ($1.96\sigma$)
and a tighter RMSE threshold of~1.70.
\begin{Verbatim}[fontsize=\scriptsize, breaklines, breakanywhere, commandchars=\\\{\}]
  # --- thresholds ---
  rmse_threshold = 1.70                          # <-- 0.05 tighter
  coverage_threshold = 72.0                      #     (percent scale)

  # --- RMSE ---
  rmse = np.sqrt(np.mean((y_true - y_pred) ** 2))

  # --- uncertainty calibration: 95% CI ---
  ci_lower = y_pred - 1.96 * y_std              # <-- 1.96*sigma CI
  ci_upper = y_pred + 1.96 * y_std
  coverage = np.mean(                            # <-- P(y_true in CI)
      (y_true >= ci_lower) & (y_true <= ci_upper)) * 100

  success = (rmse <= rmse_threshold
             and coverage >= coverage_threshold)
\end{Verbatim}

\end{tcolorbox}


\begin{tcolorbox}[
  enhanced,
  breakable,
  colback=blue!3!white,
  colframe=blue!50!black,
  title={\textbf{Example 2}},
  fonttitle=\bfseries\small,
  fontupper=\scriptsize,
  left=4pt, right=4pt, top=4pt, bottom=4pt,
  boxrule=0.5pt
]
\label{ex:eval-polymer}

\textbf{Task.}
Train a graph convolutional network to predict polymer
crystallization tendency from molecular SMILES, evaluated on a
held-out test set.

\medskip
\textbf{Key differences.}
The gold script uses a single MSE~$\le 350$ gate
(equivalent to RMSE~$\approx 18.7$).
The silver script replaces this with three complementary metrics:
MAE~$\le 25$, RMSE~$\le 30$, and $R^2 \ge 0.1$---all must pass.
The added $R^2$ gate catches degenerate models that predict near the
mean: such a model could have MSE~$\approx 300$ and pass the gold
script, but would fail the silver one due to low variance explained.

\medskip
\textbf{Gold evaluation script.}
Uses a single MSE gate with a generous threshold of~350.
\begin{verbatim}
  MSE_THRESHOLD = 350.0                          # <-- single gate

  y_true = merged["value"].values.astype(float)
  y_pred = merged["predicted_crystallization_tendency"] \
               .values.astype(float)

  mse = float(np.mean((y_true - y_pred) ** 2))

  if mse > MSE_THRESHOLD:                        # <-- only check
      return False, f"MSE too high: {mse:.4f}"
  return True, f"MSE={mse:.4f}"
\end{verbatim}

\medskip
\textbf{Silver evaluation script.}
Replaces the single MSE gate with three complementary metrics:
MAE, RMSE, and~$R^2$.
\begin{verbatim}
  MAE_THRESHOLD  = 25.0                          # <-- triple gate
  RMSE_THRESHOLD = 30.0
  R2_THRESHOLD   = 0.1

  mae  = mean_absolute_error(y_true, y_pred)
  rmse = np.sqrt(mean_squared_error(y_true, y_pred))
  r2   = r2_score(y_true, y_pred)

  all_pass = (mae  <= MAE_THRESHOLD              # <-- all three
              and rmse <= RMSE_THRESHOLD          #     must pass
              and r2   >= R2_THRESHOLD)
\end{verbatim}

\end{tcolorbox}
\subsection{Task Difficulty}
\label{app:task-difficulty}
Table~\ref{tab:solvability} details the difficulty study summarized in
\S\ref{sec:env}. We report for each model the fraction of the 75 sampled
tasks whose generated program executes without errors (execution success)
and additionally passes the silver evaluation script (evaluation success).
The gap between the two columns shows that producing runnable code is not
the main bottleneck, as most programs that execute still fail the
task-specific criteria.
\begin{table*}[h]
\centering
\begin{tabular}{lcc}
\hline
\textbf{Model} & \textbf{Evaluation Success} & \textbf{Execution Success} \\
\hline
GPT 5.4               & 38.66\% & 82.66\% \\
Claude Opus 4.6       & 36.00\% & 78.66\% \\
Qwen3 235B A22B       & 29.33\% & 61.33\% \\
Qwen3 32B             & 16.00\% & 36.00\% \\
\hline
\end{tabular}
\caption{Evaluation and execution success of different models on a subset of 75 tasks from \env, showing the non-triviality of the tasks and the room for improvement.}
\label{tab:solvability}
\end{table*}
\section{Training Details}
\label{app:training-details}
\paragraph{Training Data Collection.}
For each model, we sample 16 trajectories per task on all tasks in \env. Each trajectory contains the model's full reasoning trace and final solution, and is executed against the corresponding silver evaluation script in the environment. We retain only trajectories that successfully pass the silver evaluation and use them as training data. This yields 1,126 successful trajectories for Qwen3-4B, 1,306 for Qwen3-8B, 1,788 for Qwen3-14B, and 2,153 for Qwen3-32B. In \textbf{RFT-Distill}, student models are fine-tuned on the 2,153 successful trajectories generated by Qwen3-32B. In \textbf{RFT-Self}, each model is fine-tuned on its own successful trajectories. For Qwen3-4B-Instruct, we only apply RFT-Distill and use the final solutions without reasoning traces.

\paragraph{Supervised Fine-tuning.}
We perform fine-tuning using the LlamaFactory library \citep{zheng2024llamafactory}. We use the Qwen template and train with LoRA rank 16, LoRA alpha 32, and dropout 0.05. Training is performed in the standard supervised fine-tuning (SFT) stage over the retained successful trajectories.

\paragraph{Training Infrastructure.}
Training is conducted on NVIDIA H100 GPUs. For smaller models (4B/8B/14B), we use 4 H100 80GB GPUs, while the 32B model is trained on 8 H100 96GB GPUs. We use a learning rate of 5e-5, cosine learning rate scheduling, 1 training epoch, warmup ratio 0.05, bf16 precision, gradient checkpointing, and max gradient norm 1.0. We set both the training cutoff length and maximum sequence length to 32{,}768 tokens. Training uses per-device batch size 1 and gradient accumulation steps 1. We use distributed data parallel training for all runs.

\paragraph{Inference and Evaluation.}
We use vLLM \citep{kwon2023efficient} to serve all models for both trajectory collection and evaluation. Unless otherwise specified, we use temperature 0.2, top\_p 0.95, and max\_tokens 16k. For each evaluation setting, we run 3 independent inference runs and report the average performance across runs.



\section{ScienceAgentBench-Verified}
\label{app:sab_verified}
In this work, we introduce \textbf{ScienceAgentBench-Verified}, a manually revised version of ScienceAgentBench (SAB) that corrects a set of errors and ambiguities in the original benchmark. The issues were flagged using BenchGuard \citep{tu2026benchguardguardsbenchmarksautomated} and then checked and fixed by the authors. The revisions cover task instructions, dataset/environment files, gold programs, and gold results. In total, the verified update revises 9 task instructions (Tasks 9, 12, 26, 29, 31, 34, 35, 67, and 92) and updates gold programs and/or gold results for 3 tasks (Tasks 21, 32, and 78). These changes fix issues such as mismatched output specifications, incorrect file references, ambiguous column names, and inconsistencies between the written instruction and the implemented evaluator.

\paragraph{Summary of benchmark revisions.}
The main revisions in SAB-Verified are as follows:
(i) instruction fixes for Tasks 9, 12, 26, 29, 31, 34, 35, 67, and 92; and
(ii) gold-program and/or gold-result updates for Tasks 21, 32, and 78.
Representative example revisions include correcting the requested statistic in Task 9, changing the expected output from SMILES to drug names in Task 12, correcting input file names in Tasks 29, 34, and 35, refining the expected output schema in Task 67, and fixing matrix factor definitions in Task 92.

\paragraph{Impact of the benchmark revision.}
To validate the effectiveness of our verification, we compare several representative frontier models and coding agents on both the original SAB and SAB-Verified. On the original SAB, the strongest success rate among the evaluated settings is achieved by \textbf{Claude Code (Claude-Sonnet-4.6)} at \textbf{44.1 SR}. Replacing SAB with SAB-Verified while keeping the environment unchanged generally leads to small but non-negligible shifts in performance, often around 1--2 solved tasks. For example, Claude-Sonnet-4.6 with self-debug improves from 35.3 SR / 82.4 VER to 38.2 SR / 86.3 VER, Claude Code (Claude-Sonnet-4.6) improves from 44.1 SR / 79.4 VER to 45.1 SR / 80.4 VER, and Codex CLI (GPT-5.2) improves from 43.1 SR / 89.2 VER to 45.1 SR / 87.3 VER. After additionally incorporating the affected gold-program and gold-result fixes, the resulting performance changes are generally marginal, typically within about one solved task. This suggests that the verified benchmark mainly improves evaluation fidelity by reducing noise and ambiguity, while largely preserving the overall difficulty of the benchmark.
\section{Autoresearch Experimental Details}
\label{app:autoresearch}

This section provides the harness configuration and the full task specifications for the two Autoresearch case studies presented in \autoref{sec:autoresearch}. We select two \env tasks that naturally fit the iterative optimization setting: a classical geostatistical task with a structured objective, and a more open-ended graph-learning task requiring architecture and training choices.

\paragraph{Harness setup.} We adopt Karpathy's Autoresearch harness \citep{karpathy_autoresearch}, with minor changes to support Codex \citep{openai2025codex} and persistent memory across sessions. The original harness assumes a continuous agent session. Codex sessions are more naturally interactive, so we add a lightweight loop mechanism that allows the experiment to continue across sessions until a fixed iteration budget is reached. We also provide the agent with a persistent \texttt{result.tsv} file that records previous attempts, scores, and short descriptions, and instruct the agent to read this file before making further changes. This file serves as the agent's memory across iterations.

Each environment contains a \texttt{problem.md} file with the task instruction and rules, a \texttt{data/} directory with the task inputs, and an \texttt{evaluation.py} script adapted from the corresponding silver evaluation script. For the experiments here, we modify the evaluation scripts to return the raw optimization score rather than only a binary pass/fail verdict. We use Codex with GPT-5.4 for 40 iterations per task. The first iteration creates a baseline solution from scratch, and the remaining iterations attempt to improve it using the feedback from the evaluator.
\begin{tcolorbox}[
  enhanced,
  breakable,
  colback=blue!3!white,
  colframe=blue!50!black,
  title={\textbf{Autoresearch Task 1: Spatial Interpolation of German Temperature Data}},
  fonttitle=\bfseries\small,
  fontupper=\scriptsize,
  left=4pt, right=4pt, top=4pt, bottom=4pt,
  boxrule=0.5pt
 ]
\begin{itemize}[leftmargin=*,itemsep=1pt]
  \item \textbf{Task.} Perform Universal Kriging on German DWD station temperatures using a north-south latitude drift. The inputs are \texttt{temp\_obs.txt} (station id, latitude, longitude, temperature) and \texttt{de\_borders.txt} (German border coordinates). Fit a Spherical variogram, interpolate a $92\times112$ Germany-wide grid ($47$--$56.1^{\circ}$N, $5$--$16.1^{\circ}$E, $0.1^{\circ}$ resolution), and save the variogram plot, temperature field, mean drift field, temperature maps, cross-section plot, and model parameters.
  \item \textbf{Optimization score.} Temperature-field RMSE against the reference output; lower is better. The original binary success threshold is RMSE $<1.5$.
\end{itemize}
\end{tcolorbox}

\begin{tcolorbox}[
  enhanced,
  breakable,
  colback=blue!3!white,
  colframe=blue!50!black,
  title={\textbf{Autoresearch Task 2: Molecular Property Prediction with ChebyNet}},
  fonttitle=\bfseries\small,
  fontupper=\scriptsize,
  left=4pt, right=4pt, top=4pt, bottom=4pt,
  boxrule=0.5pt
]
\begin{itemize}[leftmargin=*,itemsep=1pt]
  \item \textbf{Task.} Train a ChebyNet graph neural network on the Tencent Alchemy PyTorch Geometric dataset. Molecules are represented as graphs with atom/node features and bond/edge structure, and the model should use Chebyshev convolutions with polynomial order $5$. Train on the development split, predict $12$ molecular properties for each validation molecule, and write \texttt{pred\_results/pred\_chebynet.csv} with \texttt{gdb\_idx} plus \texttt{property\_0}--\texttt{property\_11}.
  \item \textbf{Optimization score.} Mean absolute error averaged across all $12$ properties after merging predictions with gold outputs by \texttt{gdb\_idx}; lower is better. The original binary success threshold is MAE $<2.0$.
\end{itemize}
\end{tcolorbox}

\section{Repository Licenses}
\label{app:licenses}
We ensure that all repositories from which \env tasks are adapted are under permissive licenses. We give the full breakdown of licenses in \autoref{tab:licenses} in addition to the names of the two repositories that are under custom licenses. There are also 39 repositories that do not provide any license information. We assume that they permit use for research purposes. 
\begin{table}[t]
\centering
\small
\begin{tabular}{@{}llc@{}}
\toprule
\textbf{License} & \textbf{Repository} & \textbf{Count} \\
\midrule
MIT & & 99 \\
GNU (GPL, AGPL, LGPL) & & 43 \\
None & & 39 \\
BSD & & 29 \\
Apache & & 22 \\
CC & & 4 \\
ISC & & 1 \\
\midrule
\multicolumn{2}{@{}l}{Custom} & 2 \\
& BrainIAC & \\
& DeepDelta & \\
\midrule
\textbf{Total} & & \textbf{239} \\
\bottomrule
\end{tabular}
\caption{Distribution of repository licenses across all source repositories.}
\label{tab:licenses}
\end{table}
\section{Disclosure of AI Use}
\label{app:ai_use}
The authors have made use of AI tools for coding assistance and writing refinement. 
\end{document}